\documentclass[runningheads,orivec]{llncs}

\usepackage[T1]{fontenc}
\usepackage{subcaption}

\usepackage{graphicx}
\usepackage{booktabs}
\usepackage{amsmath}
\usepackage{hyperref}
\usepackage{amssymb} 

\usepackage{multirow}
\usepackage[table,xcdraw]{xcolor}
\usepackage{colortbl}

\begin{document}

\title{3D CT-Based Coronary Calcium Assessment: A Feature-Driven Machine Learning Framework}
\titlerunning{CT-Based  Calcium Assessment via Feature-Driven ML Framework}

\author{
  Ayman Abaid\inst{1,2}\textsuperscript{*} \and
  Gianpiero Guidone\inst{1,2}\textsuperscript{*} \and
  Sara Alsubai\inst{3,6} \and
  Foziyah Alquahtani\inst{3,6} \and
  Talha Iqbal\inst{1,2} \and
  Ruth Sharif\inst{3,6} \and
  Hesham Elzomor\inst{3,6} \and
  Emiliano Bianchini\inst{3,4,6} \and
  Naeif Almagal\inst{3,4,6,7} \and
  Michael G. Madden\inst{1,2} \and
  Faisal Sharif\inst{3,4,5,6}\textsuperscript{$\dagger$} \and
  Ihsan Ullah\inst{1,2}\textsuperscript{$\dagger$} 
}

\authorrunning{A. Abaid et al.}
%
\institute{
 Visual Intelligence Lab, School of Computer Science, University of Galway, Ireland \\
 \email{ihsan.ullah@universityofgalway.ie}
 \and
 Data Science Institute, University of Galway, Ireland
\and
 Sharif Cardiovascular Research Group, University of Galway, Ireland
\and
Department of Cardiology, Galway University Hospital, Galway, Ireland
\and
CRFG, University of Galway, Galway, Ireland
\and
 CURAM, University of Galway, Galway, Ireland
\and
Cardiology Department, NEOM Hospital, Saudi Arabia}

\maketitle    

\begin{center}
  \textsuperscript{*} Ayman Abaid and Gianpiero Guidone are the combined first authors.\\
  \textsuperscript{$\dagger$} Faisal Sharif and Ihsan Ullah are the combined senior authors.
  
\end{center}
\begin{abstract}
Coronary artery calcium (CAC) scoring plays a crucial role in the early detection and risk stratification of coronary artery disease (CAD). In this study, we focus on non-contrast coronary computed tomography angiography (CCTA) scans, which are commonly used for early calcification detection in clinical settings. To address the challenge of limited annotated data, we propose a radiomics-based pipeline that leverages pseudo-labeling to generate training labels, thereby eliminating the need for expert-defined segmentations. 
Additionally, we explore the use of pretrained foundation models, specifically CT-FM and RadImageNet, to extract image features, which are then used with traditional classifiers. We compare the performance of these deep learning features with that of radiomics features. Evaluation is conducted on a clinical CCTA dataset comprising 182 patients, where individuals are classified into two groups: zero versus non-zero calcium scores. We further investigate the impact of training on non-contrast datasets versus combined contrast and non-contrast datasets, with testing performed only on non-contrast scans. Results show that radiomics-based models significantly outperform CNN-derived embeddings from foundation models (achieving ~84\% accuracy and \textit{p<0.05}), despite the unavailability of expert annotations.

\keywords{Coronary artery disease \and Radiomic features \and Predictive models \and Machine Learning \and Agatston score
\and Computer-aided diagnosis}
\end{abstract}

\section{Introduction}
Cardiovascular diseases (CVDs) are the leading global cause of mortality, with coronary artery disease (CAD) being the most prevalent and fatal form \cite{di2024heart,eng2021automated}. While traditional CAD diagnosis relies on clinical evaluation and methods such as electrocardiography and echocardiography, coronary computed tomography angiography (CCTA) has become widely adopted due to its ability to provide high-resolution 3D visualization of coronary anatomy and plaque characteristics, enabling more precise CAD assessment. These scans are often accompanied by structured clinical reports that integrate patient data, imaging parameters, and expert interpretations \cite{kiricsli2013standardized}, guiding risk stratification and treatment planning. Coronary artery calcium (CAC) scoring is a key marker for subclinical atherosclerosis and plays a central role in risk stratification \cite{greenland2004coronary,khan2023coronary,budoff2018ten}. CAC is typically quantified manually using semi-automated software, with radiographers visually confirming coronary calcification slice-by-slice \cite{jennings2021national}. While effective, this manual process is time-consuming and operator-dependent, motivating efforts toward automation and AI-assisted analysis.
Recent advances in medical artificial intelligence (AI), particularly radiomics and deep learning, have shown considerable promise in automating and improving CAC assessment from CCTA images. In recent years, the development of sophisticated deep learning algorithms has enabled detailed analysis of coronary arteries, leading to a surge of research focused on predicting and classifying CAD. A common deep learning strategy involves using convolutional neural networks (CNNs) to extract high-level imaging features, which are then used for downstream tasks such as CAC score classification \cite{eng2021automated,ihdayhid2023evaluation,hong2022automated,zreik2018recurrent,liang2024automated}. Other objectives include plaque quantification and characterization \cite{narula2024prospective}, as well as coronary artery stenosis assessment \cite{ihdayhid2024coronary,adolf2023convolutional}. Most studies adopt a two-step pipeline: segmentation of the coronary arteries followed by feature analysis. While some approaches incorporate deep learning for automated segmentation \cite{eng2021automated,narula2024prospective,liang2024automated}, others rely on dedicated medical software for vessel-level or volume-of-interest (VOI) segmentation \cite{militello2023ct,adolf2023convolutional,park2023novel}. However, such segmentation-dependent pipelines can be difficult to scale in real-world clinical settings due to the lack of manual annotations and limited access to robust automated segmentation tools.\\
In this study, we focus on non-contrast CCTA scans, which are routinely used in clinical practice for the early detection of CAD. We also investigate how model performance varies when trained on non-contrast scans alone versus a mixed dataset of contrast and non-contrast scans, with evaluation performed on non-contrast scans only. To address the challenge of limited annotations, we propose a radiomics-based pipeline that bypasses the need for expert-defined segmentations by leveraging a pseudo-labeling strategy to generate training labels for CAD classification. We further hypothesize that foundation models trained on medical imaging, such as CT-FM \cite{pai2025vision} and RadImageNet \cite{mei2022radimagenet}, may offer complementary global features for this task. To address the challenges posed by limited data and the absence of annotations in CCTA imaging, this study explores and compares multiple feature extraction strategies to evaluate their effectiveness for calcium score classification. Specifically, we investigate whether radiomic features extracted from pseudo-labeled CCTA volumes can effectively predict calcium scores, and examine the impact of different feature extraction pipelines, including radiomics, CT-FM, and RadImageNet, on model performance in classifying CCTA volumes.

\begin{table}[ht]
\centering
\caption{Summary of Dataset}
\label{tab:summary_dataset}

\begin{tabular}{@{}p{6cm}r@{}}
\toprule
\textbf{Parameter} & \textbf{Value} \\ \midrule
Total patients & 188 \\
Average slice thickness (mm) & 0.50 \\
Image shape (Height x Width x Slices) & $512 \times 512 \times (26\text{–}2436)$ \\
Typical voxel size (mm\textsuperscript{3}) & $0.49 \times 0.49 \times 1.41$ \\
\textit{No Calcification} patients & 94 \\
\textit{Calcification} patients & 88 \\
\bottomrule
\end{tabular}
\end{table}

\section{Dataset Description}\label{dataset}
\begin{figure}[h!]
  \centering
   \includegraphics[width=0.8\textwidth]{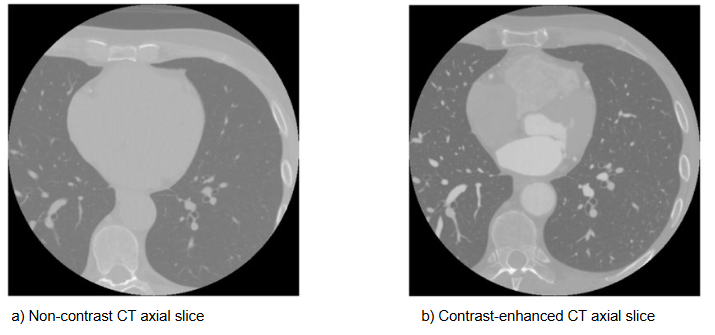} 
   \caption{Examples of axial slices from (a) contrast-enhanced and (b) non-contrast CT volumes.}
  
   \label{fig:contrast_vs_noncontrast}
 \end{figure}

In this study, due to the challenges associated with acquiring a large number of homogeneous contrast and non-contrast anonymized CT scans with reported CAC scores, and the limited availability of such datasets in the public domain, we utilized a comprehensive internal dataset collected at the University of Galway using a Research Ireland-funded GE CT scanner as part of the ACTION registry, with ethical approval (C.A. 2792). The dataset comprises 188 patients who underwent ECG-gated scan to obtain CCTA, with imaging predominantly performed using a GE Medical Systems Revolution Apex scanner. Clinical reports for 185 patients were prepared by experienced physicians following a standardized reporting protocol.
Coronary Artery Calcium (CAC) scores, typically reported as the Agatston score \cite{AgstatonScore}, are available for 182 patients. These patients constitute the final analysis cohort, with 94 labeled as \textit{No Calcification} (CAC = 0) and 88 labeled as \textit{Calcification} (CAC > 0). The primary focus of this study is the classification of patients based on their CAC status. All scan images are provided in Digital Imaging and Communications in Medicine (DICOM) format. Each file underwent anonymization checks to ensure compliance with data privacy regulations. Metadata were extracted and archived, and each imaging series was converted into Neuroimaging Informatics Technology Initiative (NIfTI) volumes to facilitate further analysis. A summary of the dataset is presented in Table~\ref{tab:summary_dataset}, and representative axial slices from contrast-enhanced and non-contrast CT volumes are shown in Figure~\ref{fig:contrast_vs_noncontrast}.

\section{Methodology}

This section outlines the feature extraction pipelines evaluated in our study and the classifiers used for the prediction of calcium score categories.

\subsection{Feature Extraction}
To address the challenge of limited labeled data, we investigated three complementary strategies for extracting high-dimensional features from the full CCTA volumes: radiomics, unsupervised representations via CT-FM, and deep features from a pretrained 2D RadImageNet model.

\subsubsection{Radiomics:}
\begin{figure}
\includegraphics[width=\textwidth]{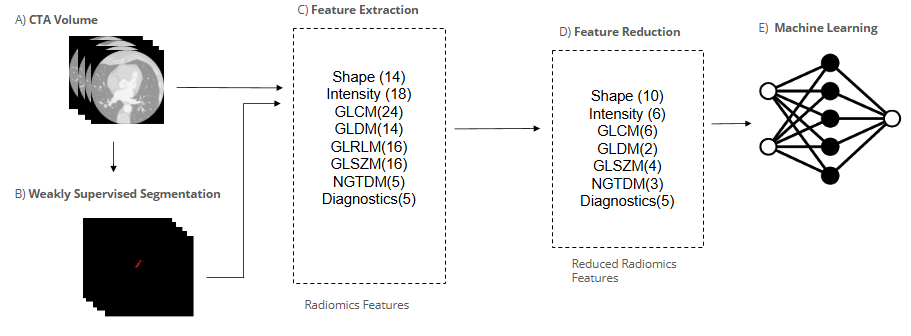}
\caption{Overview of the radiomics-based feature extraction pipeline.} 
\label{radiomics_pipeline}
\end{figure}

To overcome the lack of manual segmentations for coronary arteries, we introduce a weakly supervised segmentation approach using TotalSegmentator \cite{wasserthal2023totalsegmentator}. This tool automatically segments major anatomical structures from full-body CT scans, including the left and right coronary arteries, without requiring human-annotated training data. When provided with a full CCTA volume, TotalSegmentator generates pseudo-segmentations that serve as region-of-interest (ROI) masks for feature extraction. This weakly supervised method enables us to focus on anatomically relevant areas while avoiding noisy background information, representing a key innovation of our proposed radiomics pipeline. Using the pseudo-segmented coronary artery regions, we extracted radiomics features from the 3D volumes using the PyRadiomics library. A total of 112 features were extracted across seven categories: 14 shape features, 18 first-order intensity features, 24 gray-level co-occurrence matrix (GLCM) features, 14 gray-level dependence matrix (GLDM) features, 16 gray-level run length matrix (GLRLM) features, 16 gray-level size zone matrix (GLSZM) features, and 5 neighboring gray tone difference matrix (NGTDM) features. To reduce the dimensionality of the feature space and mitigate multicollinearity, a filter-based feature selection approach was applied using correlation thresholding to identify and eliminate highly correlated features. Following this dimensionality reduction step, the final radiomics feature set comprised 36 distinct features. An overview of the radiomics-based feature extraction pipeline is presented in Figure~\ref{radiomics_pipeline}.

\subsubsection{CT-FM:}
CT-FM \cite{pai2025vision} Feature Extractor is a large-scale, 3D image-based pretrained model specifically developed for the radiological feature extraction task. It was pretrained on a massive dataset of 146,000 CT scans obtained from the Imaging Data Commons using label-agnostic contrastive learning. For each preprocessed 3D volume, the CT-FM Feature Extractor produces a fixed number of 512 features. While feature selection was explored, it did not yield any significant improvement in the prediction performance; thus, all features were retained for subsequent analysis. The architecture is based on SegResNet, a 3D U-Net variant that employs a CNN-based encoder-decoder structure.

\subsubsection{RadImageNet:}
The RadImageNet \cite{mei2022radimagenet} feature extractor is based on a 2D ResNet50 CNN pretrained on 1.35 million annotated medical images, a large-scale medical imaging dataset designed to provide domain-specific pretraining for radiological tasks across 165 radiologic labels. Each CT scan was processed slice by slice: axial slices were preprocessed, passed through the model, and their features averaged to produce a fixed-length volumetric embedding. This domain-specific pretraining enables robust feature extraction from CT images for radiological applications.

\subsection{Prediction Models}
\label{Classifiers}
To assess the predictive capability of the extracted features, we employed a diverse set of classical machine learning algorithms, encompassing tree-based models, kernel methods, and neural networks. Specifically, we evaluated the performance of Support Vector Machines (SVM) \cite{SVM}, Random Forest \cite{randomforest}, XGBoost \cite{xgboost}, LightGBM \cite{lightgbm}, and a Multi-Layer Perceptron (MLP) \cite{hinton1990connectionist}. Hyperparameter tuning was performed for this models to classify CCTA volume-level features into one of two categories: zero or non-zero calcium score. Five-fold cross-validation was employed to determine the optimal hyperparameters. For each model, the best-performing hyperparameters were selected, and the model was subsequently evaluated on an independent non-contrast test set. Hyperparameter optimization was performed via grid search and tailored to each model architecture. 
Model performance was comprehensively evaluated using multiple metrics, including balanced accuracy, sensitivity, specificity, precision, F1-score, and negative predictive value (NPV), providing a robust assessment of classification effectiveness.
\begin{table}[t!]
\centering
\caption{Classification performance on the test set of radiomics-based models trained on datasets including both contrast-enhanced and non-contrast CT scans.}
\resizebox{\columnwidth}{!}{%
\begin{tabular}{l|l|llllll}
\hline
\textbf{Training Data} & \textbf{Model} & \textbf{Accuracy} & \textbf{Sensitivity} & \textbf{Specificity} & \textbf{PPV} & \textbf{F1-Score} & \textbf{NPV} \\ \hline

\multirow{5}{*}{\textbf{Contrast+Non-Contrast}} 
& XGBoost    & 0.81 & 0.81 & 0.81 & 0.78 & 0.84 & 0.87 \\
& SVM      & 0.75 & 0.95 & 0.56 & 0.70 & 0.81 & 0.91 \\
& Random Forest & \textbf{0.84} & \textbf{0.95} & \textbf{0.72} & \textbf{0.79} & \textbf{0.86} & \textbf{0.93} \\
& LightGBM   & 0.78 & 0.95 & 0.61 & 0.73 & 0.83 & 0.92 \\
& MLP      & 0.76 & 0.76 & 0.76 & 0.76 & 0.78 & 0.76 \\ \hline

\multirow{5}{*}{\textbf{Non-Contrast}} 
& XGBoost    & 0.79 & 0.79 & 0.79 & 0.77 & 0.81 & 0.81 \\
& SVM      & 0.81 & 1.00 & 0.61 & 0.74 & 0.85 & 1.00 \\
& Random Forest & 0.84 & 0.90 & 0.78 & 0.82 & 0.86 & 0.88 \\
& LightGBM   & \textbf{0.84} & \textbf{0.95} & \textbf{0.72} & \textbf{0.79} & \textbf{0.86} & \textbf{0.93} \\
& MLP      & 0.78 & 0.78 & 0.78 & 0.75 & 0.82 & 0.86 \\ \hline

\end{tabular}%
}
\label{tab:radiomics}
\end{table}

\begin{table}[t!]
\centering
\caption{Classification metrics on the test set by features extracted from the \\CT-FM model trained on contrast-enhanced and non-contrast CT scans.}

\centering

\label{tab:my-table}
\resizebox{\columnwidth}{!}{%
\begin{tabular}{l|l|llllll}
\hline
\textbf{Training Data} &
 \textbf{Model} &
 \textbf{Accuracy} &
 \textbf{Sensitivity} &
 \textbf{Specificity} &
 \textbf{PPV} &
 \textbf{F1-Score} &
 \textbf{NPV} \\ \hline
\multicolumn{1}{c|}{\multirow{5}{*}{\textbf{Contrast+Non-Contrast}}} &
 XGBoost &
 \textbf{0.72} &
 \textbf{0.63} &
 \textbf{0.89} &
 \textbf{0.74} &
 \textbf{0.82} &
 \textbf{0.47} \\
\multicolumn{1}{c|}{}         & SVM      & 0.52 & 0.41 & 0.73 & 0.52 & 0.64 & 0.30 \\
\multicolumn{1}{c|}{}         & Random Forest & 0.57 & 0.41 & 0.79 & 0.54 & 0.73 & 0.33 \\
\multicolumn{1}{c|}{}         & LightGBM   & 0.65 & 0.48 & 0.87 & 0.62 & 0.82 & 0.39 \\
\multicolumn{1}{c|}{}         & MLP      & 0.71 & 0.59 & 0.89 & 0.71 & 0.82 & 0.45 \\ \hline
\multirow{5}{*}{\textbf{Non-Contrast}} & XGB      & 0.61 & 0.59 & 0.80 & 0.68 & 0.64 & 0.39 \\
                    & SVM      & \textbf{0.74} & \textbf{0.67} & \textbf{0.90} & \textbf{0.77} & \textbf{0.82} & \textbf{0.50} \\
                    & Random Forest & 0.63 & 0.44 & 0.86 & 0.59 & 0.82 & 0.38 \\
                    & LightGBM   & 0.66 & 0.59 & 0.84 & 0.70 & 0.73 & 0.42 \\
                    & MLP      & 0.69 & 0.56 & 0.88 & 0.68 & 0.82 & 0.43 \\ \hline
\end{tabular}%
}

\label{tab:CTFMresults}

\end{table}

\begin{table}[t!]
\centering
\caption{Classification metrics on the test set by features extracted from the RadImageNet model trained on contrast-enhanced and non-contrast CT scans.}

\resizebox{\columnwidth}{!}{%
\begin{tabular}{@{}l|l|llllll@{}}
\hline
\textbf{Training Data} &
 \textbf{Model} &
 \textbf{Accuracy} &
 \textbf{Sensitivity} &
 \textbf{Specificity} &
 \textbf{PPV} &
 \textbf{F1-Score} &
 \textbf{NPV} \\
\hline
\multirow{5}{*}{\textbf{Contrast+Non-Contrast}}
 & XGBoost    & \textbf{0.60} & \textbf{0.60} & \textbf{0.60} & \textbf{0.60} & \textbf{0.55} & \textbf{0.61} \\
 & SVM      & 0.57 & 0.74 & 0.41 & 0.54 & 0.62 & 0.63 \\
 & Random Forest & 0.60 & 0.56 & 0.65 & 0.59 & 0.57 & 0.62 \\
 & LightGBM   & 0.55 & 0.50 & 0.60 & 0.53 & 0.51 & 0.57 \\
 & MLP      & 0.53 & 0.53 & 0.53 & 0.50 & 0.50 & 0.55 \\
\hline
\multirow{5}{*}{ \textbf{Non-Contrast}}
 & XGBoost    & 0.59 & 0.59 & 0.59 & 0.64 & 0.48 & 0.59 \\
 & SVM      & \textbf{0.60} & \textbf{0.50} & \textbf{0.70} & \textbf{0.60} & \textbf{0.55} & \textbf{0.61} \\
 & Random Forest & 0.60 & 0.32 & 0.89 & 0.73 & 0.44 & 0.59 \\
 & LightGBM   & 0.63 & 0.61 & 0.65 & 0.61 & 0.61 & 0.65 \\
 & MLP      & 0.54 & 0.54 & 0.54 & 0.56 & 0.37 & 0.55 \\
\hline
\end{tabular}%
}
\label{tab:radimagenet}
\end{table}

\section{Experiments and Results}
All experiments were conducted on an NVIDIA RTX 4500 Ada Generator using PyTorch. Radiomics features were extracted with PyRadiomics, and deep learning embeddings were obtained from pretrained models (CT-FM and RadImageNet) without fine-tuning.

\subsection{Preprocessing and Setup}

For all approaches, DICOM slices were first reconstructed into 3D volumes to enable volumetric analysis. For Radiomics approach, volumes without detectable coronary arteries identified using a segmentation model were excluded from further analysis. Radiomics features were then extracted from the remaining volumes (181 when the training data consisted of only non-contrast scans, and 485 when the training data included both non-contrast and contrast scans)  using the PyRadiomics library. These features served as inputs to subsequent feature reduction and classification models. For the CT-FM approach, the 3D volumes underwent several preprocessing steps: they were reoriented to the standard Superior Posterior Left (SPL) anatomical coordinate system, intensity values were clipped to [$-$1024, 2048] range, linearly scaled to [0, 1], and background regions were removed to minimize computational overhead. In contrast, the RadImageNet pretrained model extracted features slice-by-slice from axial images, with the final volume-level embedding obtained by averaging the embeddings across all slices. Both CT-FM and RadImageNet generated fixed-length embeddings, which were subsequently used as inputs for the classification tasks.
In clinical practice, non-contrast CT scans are standard for calcium scoring. To assess the effect of contrast enhancement, we prepared two datasets: (1) a mixed dataset with contrast and non-contrast scans for training (80\%) and non-contrast scans for testing (20\%), and (2) a non-contrast-only dataset for both training (80\%) and testing (20\%). Training datasets were used for hyperparameters tuning via grid search and classifier performance were evaluated on the test datasets.

\subsection{Results}
We evaluated the effectiveness of radiomics and deep learning-based features for binary calcium score classification using both contrast-enhanced and non-contrast CT scans, separately and in combination. Radiomics-based models (Table \ref{tab:radiomics}) showed the best overall performance, with Random Forest achieving the highest accuracy of 84\%, sensitivity of 95\%, and specificity of 72\%. XGBoost and LightGBM also performed well with accuracies of 81\% and 78\%, respectively. Models using CT-FM features (Table \ref{tab:CTFMresults}) showed lower performance, with the MLP model reaching an accuracy of 74\% and F1-score of 82\% on the non-contrast dataset. Sensitivity was moderate across models, decreasing further when trained only on non-contrast data. Classification using RadImageNet features (Table \ref{tab:radimagenet}) resulted in the lowest accuracy, with the best model (LightGBM) reaching 63\% accuracy on non-contrast data. Most models achieved accuracies between 55-60\%, with modest sensitivity and specificity. Overall, radiomics features consistently outperformed learned features from CT-FM and RadImageNet extractors. Among classifiers, Random Forest and XGBoost demonstrated the most balanced and highest performance, indicating that classical radiomics features remain highly effective for calcium score categorization, while learned features require further refinement to match this performance. We also performed a statistical analysis to determine whether the performance of the radiomics model differs significantly from that of the CT-FM model. Paired \textit{t}-tests revealed statistically significant differences in both the combined contrast and non-contrast dataset (accuracy: \textit{p} = 0.033; F1-score: \textit{p} = 0.017) and the non-contrast dataset alone (accuracy: \textit{p} = 0.031; F1-score: \textit{p} = 0.024). These results further suggest that radiomics features provide a more discriminative representation for classification compared to the CT-FM approach.

\section{Conclusion}
In this study, we developed and evaluated radiomics-based and deep learning approaches for the classification of coronary artery calcium (CAC) scores using non-contrast CCTA scans. Our primary focus was on categorizing patients into zero versus non-zero calcium score groups, an important early indicator for coronary artery disease risk stratification.
We demonstrated that radiomics-based models, leveraging interpretable features, significantly outperformed CNN-based embeddings derived from foundation models such as CT-FM and RadImageNet. Radiomics features capture well-defined quantitative descriptors of image texture and intensity, contributing to their strong performance, reproducibility, and interpretability. These results highlight the continued relevance and effectiveness of radiomics, particularly in scenarios with limited annotations and heterogeneous imaging data. Importantly, our findings support the clinical utility of radiomics-based models as a reliable and explainable tool for early CAD risk assessment, especially in resource-limited settings. In contrast, models leveraging CT-FM and RadImageNet embeddings achieved lower performance, particularly on non-contrast CT scans. This suggests current deep feature extractors may lack task-specific sensitivity, and could benefit from fine-tuning or improved fusion strategies to enhance feature expressiveness and generalizability.\\
For future work, we plan to incorporate clinical report text using multimodal fusion techniques, enabling models to leverage both visual and textual features. Additionally, instead of binary classification, we aim to explore multi-class calcium scoring (e.g., 0 (no), 1–10 (minimal), 11–100 (mild), 101–400 (moderate), >400 (severe)), which would provide a more granular and clinically meaningful risk stratification. Additionally, expanding datasets and including segmentation masks may further enhance learned feature extraction.
\subsubsection{Prospect of application} 
This work enables accurate, automated calcium score classification from ECG-gated scans of non-contrast CCTA, facilitating early cardiovascular risk assessment. It integrates into clinical workflows to support radiologists, improve patient stratification, and guide preventive treatments. Requiring minimal training data and no expert annotation, it is practical for rapid deployment across diverse healthcare settings

\subsubsection{\ackname} 

This publication has emanated from research conducted with the financial support of Taighde Éireann – Research Ireland under Grant No. 18/CRT/6223, and SFI/12/RC/2289\_P2 the Insight Research Ireland Centre for Data Analytics, and SFI 17/RI/5353. The authors thank Eileen Coen and Aoife Joyce for their efforts in the data collection procedure.

\subsubsection{\discintname}
The authors have no competing interests to declare that are relevant to the content of this article.

%
%
%

%
\bibliographystyle{unsrt} 
\newpage
\bibliography{references}

\end{document}